\pdfoutput=1
\documentclass[lettersize,journal]{IEEEtran}
\usepackage{amsmath,amsfonts}
\usepackage{algorithmic}
\usepackage{algorithm}
\usepackage{array}
\usepackage[caption=false,font=normalsize,labelfont=sf,textfont=sf]{subfig}
\usepackage{textcomp}
\usepackage{stfloats}
\usepackage{url}
\usepackage{verbatim}
\usepackage{graphicx}
\usepackage[numbers]{natbib}
\usepackage{csquotes}
\usepackage{cleveref}
\usepackage{multirow}

\DeclareMathOperator*{\argmax}{arg\,max}

\newcommand{\Substack}[1]{\hbox to 0pt{\hss $\substack{ #1}$\hss}}

\usepackage{color}
\usepackage[table,cmyk]{xcolor}

\newcommand\blfootnote[1]{%
  \begingroup
  \renewcommand\thefootnote{}\footnote{#1}%
  \addtocounter{footnote}{-1}%
  \endgroup
}

\RequirePackage{filecontents}

\begin{document}

\bstctlcite{IEEEexample:BSTcontrol}
\title{Task-oriented Document-Grounded Dialog Systems by HLTPR@RWTH for DSTC9 and DSTC10}

\author{
    David Thulke,\textsuperscript{\rm 1,2}
    Nico Daheim,\textsuperscript{\rm 1,*}
    Christian Dugast,\textsuperscript{\rm 2}
    Hermann Ney\textsuperscript{\rm 1,2}\\\,\\
    \textsuperscript{\rm 1}Human Language Technology and Pattern Recognition Group, RWTH Aachen University \\
    \{thulke, ney\}@hltpr.rwth-aachen.de \\
    \textsuperscript{\rm 2}AppTek GmbH, Aachen \\
    cdugast@apptek.com \\
}


\maketitle
\blfootnote{*Now affiliated with Ubiquitous Knowledge Processing Lab, Technical University of Darmstadt}%
\begin{abstract}
This paper summarizes our contributions to the document-grounded dialog tasks at the 9th and 10th Dialog System Technology Challenges (DSTC9 and DSTC10).
In both iterations the task consists of three subtasks: first detect whether the current turn is knowledge seeking, second select a relevant knowledge document, and third generate a response grounded on the selected document.
For DSTC9 we proposed different approaches to make the selection task more efficient. The best method, Hierarchical Selection, actually improves the results compared to the original baseline and gives a speedup of 24x.
In the DSTC10 iteration of the task, the challenge was to adapt systems trained on written dialogs to perform well on noisy automatic speech recognition transcripts.
Therefore, we proposed data augmentation techniques to increase the robustness of the models as well as methods to adapt the style of generated responses to fit well into the proceeding dialog.
Additionally, we proposed a noisy channel model that allows for increasing the factuality of the generated responses.
In addition to summarizing our previous contributions, in this work, we also report on a few small improvements and reconsider the automatic evaluation metrics for the generation task which have shown a low correlation to human judgments.
\end{abstract}


\section{Introduction}
\IEEEPARstart{D}{ocument-grounding} allows incorporating unstructured information into dialog systems to make them more interesting or to extend the scope of task-oriented dialog systems.
The latter are typically restricted to information provided by an application-specific structured database.
In practice, this rarely covers all possible information needs users may have.
This information is in many cases available in unstructured documents, e.g. on the web, such as FAQ documents or reviews.
The aim of Track 1 of the 9th Dialog System Technology Challenge (DSTC9) \cite{kimDomainAPIsTaskoriented2021} and Task 2 of Track 2 of the 10th Dialog System Technology Challenge (DSTC10) \cite{kimKnowledgegroundedTaskorientedDialogue2022} 
To do so, the task at hand is split up into three subtasks, namely Knowledge-seeking \emph{Turn Detection} to identify those questions that can not be answered by an existing API, \emph{Knowledge Selection} to retrieve relevant documents, and \emph{Response Generation} to generate a suitable system response.

For the DSTC9 challenge, we focussed on two aspects. First, making the selection subtask more efficient.
Therefore, we proposed a Hierarchical Selection approach that gives a speedup of 25x while simultaneously giving better results.
We further proposed to use a bi-encoder model for the task which gives an additional 100x speedup and would be usable even with larger knowledge bases.
Second, we proposed a Retrieval Augmented Generation model for the generation subtask that can condition a response on multiple documents.

For the DSTC10 challenge, in which the task was to adapt systems to noisy ASR transcripts, we proposed multiple data augmentation techniques to adapt the system to new domains as well as to make it more robust to noisy inputs.
Further, we studied methods to adapt the style of generated responses to better fit into a spoken dialog.
We also proposed a noisy channel model that allows for generating more factual responses.

In addition to summarizing our submissions for DSTC9 and DSTC10, we apply our recent findings to the DSTC9 test set and show that we can outperform the results of all finalists in the selection and generation task.
Further, we reconsider the bi-encoder model and improve the training using recently proposed methods for contrastive learning. This allows us to reduce the gap between cross-encoder and bi-encoder models.
Additionally, we reconsider the automatic metrics for the generation task.
The results of the DSTC10 challenge have shown that the current metrics have low or even no correlation to human judgments.
As an alternative, we evaluate a set of recently proposed factuality metrics for this task and show that they have a higher correlation to human judgments.

\section{Task Description}
\label{sec:task_description}

\begin{table*}
\caption{Data statistics of the training and evaluation data splits of DSTC9 and DSTC10.} 
\label{table:data_statistics}
\begin{center}
\begin{tabular}{|l|r|r|r|r|r|r|r|} \hline
    Split      & \# dialogs & \# ks dialogs & \# documents & \# domains & \# entities \\ \hline\hline
    Train      & 72,518 & 19,184 & \multirow{2}{*}{2,900}& \multirow{2}{*}{4}& \multirow{2}{*}{145} \\ \cline{1-3} 
    Validation$_{\text{DSTC9}}$ & 9,825 & 2,673 & & &\\ \hline
    Test$_{\text{DSTC9}}$ & 4,181 & 1,981 & {12,039} & {5} &{668}\\ \hline
    Validation$_{\text{DSTC10}}$ & 263 & 104 & \multirow{2}{*}{9,139} & \multirow{2}{*}{3} &\multirow{2}{*}{523}\\ \cline{1-3}
    Test${_{\text{DSTC10}}}$ & 1,988 & 683 & & &\\ \hline
\end{tabular}
\end{center}
\end{table*}

Given a dialog context $u_1^T = (u_1, \dots, u_T)$ consisting of $T$ user and agent turns, the task in a dialog system is to generate an agent response $u_{T+1}$.
Therefore, we assume that our dialog system consists of a component that can create a response by retrieving information from relevant unstructured documents from a knowledge base $K$ and a separate component that can handle all other user requests.
In the context of the challenges, this is split into the following three subtasks:

\subsubsection{Detection} For knowledge-seeking turn detection, the system has to decide whether the current user turn $u_T$ requires unstructured knowledge access or can be handled by some other component in the dialog system.
In the latter case, it is assumed that a corresponding component exists and generates an agent response. In the former case, we continue with Subtask 2.
Formally, we want to implement the following decision rule $r_1$: \[
r_1(u_1^T, K) = \begin{cases}
					1 & \text{if }\exists k \in K \text{ s.t. } k \text{ answers } u_T \\
					0 & \text{otherwise}
				   \end{cases}
\]
\subsubsection{Selection} In Knowledge Selection, in the general case, the system has to find the documents $K^\prime$ from the knowledge base $K$ that contain information relevant to create a response to the last user turn $u_T$.
Formally, this is expressed by the following decision rule:
\[
	f_2(u_1^T, K) = \{k \mid k \in K \land k \text{ relevant to } u_1^T\} = K^\prime
\]
In the datasets for both iterations of the challenge, it is always the case that exactly one document is relevant to one turn.
This means $|K^\prime| = 1$.
\subsubsection{Generation} Finally, response generation is the task of generating an agent response $u_{T+1}$ that accurately reflects the information from the selected documents $K^\prime$ and is appropriate given the dialog context $u_1^T$.
Thus, the task can be defined as
\[
	r_3(u_1^T, K^\prime) = u_{T+1} 
\]

\subsection{Data}

The data that is provided as part of the challenges is based on the MultiWOZ 2.1 dataset \cite{budzianowskiMultiWOZLargeScaleMultiDomain2018,ericMultiWOZConsolidatedMultiDomain2020}.
MultiWOZ is a task-oriented dialog dataset consisting of 10,438 dialogs covering 7 different domains (e.g. hotels, restaurants, train) related to local information in the City of Cambridge.
Each of these domains is defined by an ontology and a database of corresponding entities.
The dialogs are all written and were collected using the Wizard-of-Oz methodology.
For the DSTC challenges, \citet{kimDomainAPIsTaskoriented2020} extended the corpus by new user turns that require information beyond the existing database and corresponding system responses.
Additionally, a knowledge base for each domain was created by collecting question-answer pairs from FAQ sections of relevant websites.
Thus, each document corresponds to a domain, an entity, and consists of a question and an answer.
In total, for the original training and validation data, 21.857 new user and agent turns and 2.900 documents were collected. 
These two splits are restricted to the four domains hotel, restaurant, taxi, and train.
The latter two do not contain any entities and corresponding documents are relevant for the whole domain.

For the test set of the DSTC9 challenge a new domain, attraction was added.
It contains 4.181 additional dialogs of which 1.981 have knowledge-seeking turns and 12.039 documents.
Around half of these dialogs are from the MultiWOZ dataset augmented as described above.
The other half are human-to-human conversations about tourist information for a new locality San Francisco.
Of these, 90\% are written conversations and 10\% transcriptions of spoken conversations.
Additionally, these dialogs only cover the hotel, restaurant, and attraction domains.
This created the challenge that systems should be able to adapt to new domains and localities.
Further, systems had to be able to able to also handle transcripts of spoken conversations in addition to written ones.

The second iteration of the challenge at DSTC10 put its focus on adapting systems for the task to spoken conversations.
To make the conditions more realistic, systems should not be evaluated on human transcripts of these conversations, but on transcripts generated by an Automatic Speech Recognition (ASR) system.
Therefore, for the validation set the organizers re-released the subset consisting of 263 spoken conversations from the DSTC9 test set transcribed using an ASR system.
For each user turn, they provide an n-best list generated by the ASR system with corresponding confidence scores.
The test set consisted of 1,988 new spoken dialogs from the San Francisco locality transcribed with the same ASR system.

\Cref{table:data_statistics} gives a more detailed overview of the different splits published in the two challenges.
For more details and examples illustrating the task, we refer readers to the original task description papers \cite{kimDomainAPIsTaskoriented2020} and \cite{kimHowRobustEvaluating2021}. 

\section{Methods}
In this section, we discuss the different methods used by us to approach this task.
For all tasks, we fine-tuned the large variants of pre-training transformer encoder models like BERT \cite{devlinBERTPretrainingDeep2019}, RoBERTa \cite{liuRoBERTaRobustlyOptimized2019}, DeBERTa \cite{he2021debertav3}, or BART \cite{lewisBARTDenoisingSequencetoSequence2020} with around 0.4B parameters.
For classification tasks, we add a classifier on top of the final layer output of the classification token (i.e. \texttt{[CLS]} for BERT and DeBERTa and \texttt{\textless s\textgreater} for RoBERTa).

Due to the inherent length limitations of these models (e.g. the trained positional embeddings) or due to memory constraints, input sequences to these models may need to be truncated to maximum length.
We typically do this by truncating the oldest utterances of the complete dialog until we reach the maximum input length.



\subsection{Detection}

As in the baseline model proposed by \citet{kimDomainAPIsTaskoriented2020}, we model knowledge detection as a binary classification task.
Therefore, we add a simple classifier consisting of two linear layers on top of the hidden state of the last layer of the classification token.
To limit the length of the input, we only pass the last three utterances of the dialog to the model and additionally truncate the input sequence if it exceeds 384 tokens.
The model is trained using cross-entropy.

\subsection{Selection}

In the selection subtask, the goal is to retrieve the most relevant document from the knowledge base for the current knowledge-seeking user turn.
Typically, this is modeled as a ranking task of the documents given the user turn.
Therefore, a relevance function $\text{sim}(u_1^T, k)$ is defined which measures the relevance of a document $k$ to a user turn and its context $u_1^T$.
In the following, we discuss a few different approaches to implement such a relevance function.

\subsubsection{Cross-Encoder}

The original baseline implementation for a relevance function for this task suggested by \citeauthor{kimDomainAPIsTaskoriented2020} is a cross-encoder.
Therefore, the dialog context and the document are concatenated and separated by a special separator token.
This sequence is then passed as input to the transformer model that is trained to predict whether the given document is relevant to the dialog or not.
During training for each sample, a positive dialog and document pair is used and a set of negative pairs.
Previous work on this task has shown that a good selection of negative samples is critical for good results \cite{heLearningSelectExternal2021, jinCanBeFurther2021}.
In our case, we randomly sample one document from a different domain, one document from the same domain but a different entity, and one document from the same entity.

A major drawback of this approach is that it requires a full forward pass of the model for each dialog context and document pair at inference time.
Already for smaller knowledge bases in the order of thousandths of documents, as used in the challenges, this becomes prohibitively expensive.

\subsubsection{Hierarchical Selection}

One way to make the selection more efficient, as we proposed for DSTC9 \cite{thulkeEfficientRetrievalAugmented2021}, is Hierarchical Selection.
Instead of calculating the relevance score for each document, the problem can be divided by first identifying the relevant domain, then the relevant entity, and finally the relevant document.
This allows to significantly reduce the search space and thus the number of required forward passes through the model.
For each step, we train a separate cross-encoder model as described above.
We consider two variants: in the first, three separate models are used to select the relevant domain, entity, and document.
In the second, one model $p_E$ is used to jointly select the entity, and domain and one model $p_D$ to select the document.
To train the first variant we sample negatives for each of the corresponding categories (i.e. for the entity model we sample negative entities within the same domain).
For the second variant, for the entity selection model, we randomly sample one entity with a different domain and two entities with the same domain as the negative samples.
For the document selection model, we sample three documents of the same entity as negative samples.

In the original variant proposed for the DSTC9 challenge, we used a greedy search method.
This means after each subselection step, we only considered the most relevant domain or entity and then only considered entity or documents for that specific domain or entity.
This causes the issue that in cases where for example the relevance score of the two best entities is relatively close, this ambiguity may be resolved by comparing the relevance of the associated documents.

For the DSTC10 challenge, we proposed an alternative beam search method for inference \cite{thulkeAdaptingDocumentGrounded2022}.
Specifically, we proposed to consider all entities with an entity relevance score \(p_E(r|e,u_1^T)\) that is within a threshold \(t \leq 1\) of the most relevant entity \(\hat e\):
\begin{align}
    \hat k = (\hat e, \hat d)= \argmax_{\Substack{k = (e, d)\\p(e \mid  u_1^T) > t \cdot p(\hat e \mid u_1^T)}} \,\, p_E(e \mid u_1^T)^\gamma \cdot p_D(d \mid e, u_1^T)
\end{align}
Additionally, a scaling factor \(\gamma\) was added allowing to control the influence of the models on the final selection.

\subsubsection{Bi-Encoder}

In cases where larger knowledge bases are used or a lot of documents for a single domain or entity are available, the hierarchical selection approach might still not achieve latencies applicable for real-time applications.
Another solution, we proposed for DSTC9 \cite{thulkeEfficientRetrievalAugmented2021}, is to use a bi-encoder architecture.
There, separate encoders are used to transform the dialog and the document into a single embedding vector.
Then a relevance score is calculated using a similarity function or distance metric to compare these fixed vectors.
This approach allows precomputing of the embedding vectors of the whole knowledge base.
At inference time it is then only necessary to do a single forward pass through the model to calculate the embedding of the dialog.
The most relevant document can then be found by doing a nearest neighbor search over the precomputed embeddings.

For the DSTC9 challenge, we used two loss functions and a corresponding distance function to train the models.
The first is the triplet loss \cite{weinberger2009distance} and the euclidean distance.
Given an anchor, in our case the encoded dialog, and a positive and negative document, the loss trains the encoders so that the distance between the anchor and positive sample is lower than the distance to a negative document by a margin \(\epsilon\).

For the second method, we use the dot product between the embeddings created by the encoder \(E\) as a similarity measure.
We train the model, given an anchor \(a\), to correctly classify a positive sample given a positive sample \(p\) and a set of negative samples \(N\).
Mathematically the loss is the negative log-likelihood of the correct positive sample:
\[
    L = - \log \frac{\exp{\left(E(a) \cdot E(p)\right)}}{\sum_{s \in N \cup \{p\}} \exp{\left(E(a) \cdot E(s)\right)}}    
\]
The anchor can either be a dialog context and the other samples are relevant and irrelevant documents or the other way around.
Negative samples are randomly sampled from the knowledge base.

In this work, we experiment with some modifications to the second loss to improve the results.
First, we follow the NT-Xent loss \cite{chen2020simple} and use the cosine similarity as a distance function and scale its value by a temperature hyperparameter.
Further, we use in-batch negatives. That means instead of separately sampling negatives for each sample, we consider all other positive samples in the batch as negatives for the current sample.

\subsection{Generation}

The generation task can be formulated as a sequence-to-sequence task.
Given the concatenation of the dialog and the selected knowledge document, the task of the model is to generate the corresponding agent response.
This model is often referred to as \emph{direct model} and can be expressed by the following probability:
\begin{align}
    \label{eq:direct_model}
p\left(w_n  \mid w_1^{n-1}, u_1^T, K^\prime \right) \text{, }
\end{align}
We use an encoder-decoder transformer model BART \cite{lewisBARTDenoisingSequencetoSequence2020} for this task.
At inference time the model is decoded using beam search.

\subsubsection{Retrieval Augmented Generation}

The baseline approach for response generation only considers the single best selected knowledge document.
In some cases, multiple documents might contain relevant information to generate a response.
Further, by making a hard decision for a single knowledge document in the selection step, we introduce errors that are propagated to the response generation.
This motivates us to reformulate our selection and generation task into a single task, i.e. to generate a response based on all of our knowledge documents.
The approach is similar to what \citet{lewisRetrievalAugmentedGenerationKnowledgeIntensive2020} propose and to other retrieval augmented models like REALM \cite{guuREALMRetrievalAugmentedLanguage2020}.
Mathematically, we can formulate this as a marginalization over the selected knowledge document \(k\) which we introduce as a latent variable:
\[
    p(u_{T+1} | u_1^T; K) = \sum_{k \in K} p(u_{T+1}, k | u_1^T; K)
\]
which can then be further split into a selection probability, i.e. the probability of a knowledge document given a dialog context, and a generation probability which corresponds to the baseline model for generation:
\[
    p(u_{T+1}, k | u_1^T; K) = \underbrace{p\left (k \mid u_1^T ; K \right )}_\textrm{selection} \cdot \overbrace{p\left (u_{T+1} \mid u_1^T, k ; K\right )}^{\textrm{generation}}
\]
The same decomposition can also be applied on the token level which allows us to use this as a drop-in replacement for our current generation probability.
To be able to calculate this efficiently during training and testing, we approximate the sum over all knowledge documents \(K\) by a sum over the top \(n\) documents.
To ensure that the model is still normalized, we renormalize the selection probabilities over this subset.
In our experiments, we typically use \(n=5\) and ensure that the correct knowledge document is always included in the top \(n\) documents during training.
For the generation probability, we use the same model as in the baseline.
In theory, this model allows us to train the selection and generation models jointly.
However, calculating the selection probabilities using the cross-encoder models during training on the fly is not feasible, even when using the Hierarchical Selection models.
Therefore we calculate these probabilities in a previous step and keep them fixed during training.

Fortunately, using the bi-encoder model, training both models jointly becomes feasible.
Therefore, we keep the knowledge document encoder fixed and only fine-tune the dialog context encoder.
The top \(n\) knowledge documents can then be effectively retrieved during training.

\subsubsection{Style Adaptation}

To maintain a fluent conversation, generated responses should be naturally connected to the context of the dialog and thus match the style of preceding utterances.
For DSTC10 this becomes a challenge since most of the training data are written dialogs but the evaluation is one spoken dialogs.
Hence, we look for methods to encourage the model to generate answers in spoken style with either no or only a few in-domain samples.
While the direct model could infer the style of the dialog already from the context $u_1^T$, we further introduce a style token as a form of explicit conditioning.
Then, the \Cref{eq:direct_model} becomes:
$$
p\left(w_n  \mid w_1^{n-1}, u_1^T, K^\prime, s \right) \text{, }
$$
where $s$ is a special $\langle \text{written} \rangle$ or $\langle \text{spoken} \rangle$ token that is added to the vocabulary.

\subsubsection{Noisy Channel Model}

Besides the style, an additional goal to increase the quality of generated responses is to increase their factuality. Thus, we try to find a model that explicitly favors faithfulness to document grounding.
We use Bayes Theorem to derive a noisy channel formulation for document-grounded response generation as follows:
\begin{align*}
    & \arg\max_{w} p(w \mid u_1^T, K') \\
    &= \arg\max_{w} p(w, u_1^T, K') \\
    &= \arg\max_{w} \underbrace{p(K' \mid w, u_1^T)}_{\text{channel model}} \cdot \underbrace{p(w \mid u_1^T)}_{\Substack{\text{response generation}\\ \text{model}}}
\end{align*}

First of all, we can see that the advantage of having an ungrounded response generation model which can be trained on large amounts of textual data in the new domain without requiring document grounding is retained.
Furthermore, the channel model now encourages that the response explains the document grounding sufficiently well which could prevent the model from leaving out important details and mitigate the explaining-away effect \cite{liuPretrainingNoisyChannel2021a}.

However, decoding the noisy channel model directly is computationally intractable.
Hence, we use two different approximate decoding methods.
First of all, we experiment with \emph{reranking} generations obtained from a proposal model, for which we use the direct model.
That is, we first decode $k$ sequences from the direct model and then obtain the final response as the highest-scoring sequence under the log-linear model combination
\begin{align}
    \label{reranking}
    \begin{split}
        \hat{w} = \arg\max_{w} \Bigl \{ & \log p\left(w \mid u_1^T, K^\prime\right)+ \\
        & \lambda_2 \cdot \log p\left(K^\prime \mid w, u_1^T\right)+ \\
        & \lambda_1 \cdot \log p\left( w \mid u_1^T\right) \Bigr \}
    \end{split}
\end{align}
We interpolate with the direct model to encourage sequences with high direct model likelihood which has proven beneficial in other tasks \cite{yuNeuralNoisyChannel2017, liuPretrainingNoisyChannel2021a}.
While comparatively efficient, the method is limited by the proposal model, since the noisy channel formulation can only re-rank an n-best list of complete sequences.
To address this we additionally proposed an \emph{online decoding} algorithm.
Based on the results of our previous work \citet{thulkeAdaptingDocumentGrounded2022,noisyChannelPaper2022}, we concluded that the advantage of online decoding is negligible and does not outweigh the additional computation time required for online decoding.

\subsection{Data Augmentation}

To adapt the model to the knowledge documents from the new domains and localities, we generate additional knowledge-seeking dialog samples based on the documents in our knowledge base.
Therefore, for each document, we randomly select one dialog from the original MultiWOZ corpus in the same domain, replace the entity in the document with an entity from the dialog, and add the questions of the (faq) document as a new knowledge-seeking turn.
This way, we add 16,675 new samples to the training data that can be used to train models for the detection and selection task.

\subsection{Adaptation to Spoken Dialogs for DSTC10}

In contrast to the written training data, the ASR transcripts in the DSTC10 validation and test data are lower-cased and do not contain punctuation.
This creates a mismatch between the training and evaluation data.
We remove this information from the written text so that it becomes more similar to the ASR transcripts.
In addition to that, we write out numbers (e.g. \emph{42} $\mapsto$ \emph{forty two}) and spell out abbreviations (e.g. \emph{mm} $\mapsto$ \emph{millimeters}).

To make use of the n-best list provided by the ASR system, we pass each ASR hypothesis to our model and experiment with two different strategies.
The \emph{best} strategy selects the highest score of all hypotheses. The \emph{weighted} strategy calculates the weighted sum of all scores based on the (renormalized) probabilities of the ASR hypotheses.
Even though the \emph{weighted} strategy is the mathematically more sound option, as it treats the ASR hypotheses as a latent variable, we observe the highest F1 scores on the validation data with the \emph{best} strategy.

\section{Metrics}

\begin{table*}
    \caption{Correlation of the automatic metrics to the human judgments for appropriateness (app.) accuracy (acc.) for~finalists~results~of~the~DSTC9~and~DSTC10~in~Spearman's $\rho$.}
    \label{tab:correlation}
    \begin{center}
    \begin{tabular}{ll|rrrrrrrr}
        \hline
        && \multicolumn{1}{l|}{detection}   & \multicolumn{1}{l|}{selection}     & \multicolumn{3}{l|}{generation}          & \multicolumn{3}{l}{factuality} \\
        &              & \multicolumn{1}{r|}{F1}          & \multicolumn{1}{r|}{R@1}  & \multicolumn{1}{r|}{BLEU-1} & \multicolumn{1}{r|}{METEOR} & \multicolumn{1}{r|}{ROUGE-L} & \multicolumn{1}{r|}{BLEU-1} & \multicolumn{1}{r|}{F1} & \multicolumn{1}{r}{$Q^2$} \\ \hline \hline
        DSTC9  & App. & \cellcolor[cmyk]{0.6454545454545455,0.32272727272727275,0,0}\textcolor{white}{0.65} & \cellcolor[cmyk]{0.9363636363636365,0.46818181818181825,0,0}\textcolor{white}{0.94} & \cellcolor[cmyk]{0.5545454545454546,0.2772727272727273,0,0}\textcolor{white}{0.55} & \cellcolor[cmyk]{0.3727272727272728,0.1863636363636364,0,0}\textcolor{white}{0.37} & \cellcolor[cmyk]{0.5454545454545455,0.27272727272727276,0,0}\textcolor{white}{0.55} & \cellcolor[cmyk]{0.30000000000000004,0.15000000000000002,0,0}\textcolor{white}{0.30} & \cellcolor[cmyk]{0.3272727272727273,0.16363636363636366,0,0}\textcolor{white}{0.33} & \cellcolor[cmyk]{0.06363636363636364,0.03181818181818182,0,0}\textcolor{black}{0.06} \\
               & Acc. &  \cellcolor[cmyk]{0.7363636363636363,0.36818181818181817,0,0}\textcolor{white}{0.74} & \cellcolor[cmyk]{0.6272727272727273,0.31363636363636366,0,0}\textcolor{white}{0.63} & \cellcolor[cmyk]{0.6363636363636364,0.3181818181818182,0,0}\textcolor{white}{0.64} & \cellcolor[cmyk]{0.7272727272727273,0.36363636363636365,0,0}\textcolor{white}{0.73} & \cellcolor[cmyk]{0.4272727272727273,0.21363636363636365,0,0}\textcolor{white}{0.43} & \cellcolor[cmyk]{0.8000000000000002,0.4000000000000001,0,0}\textcolor{white}{0.80} & \cellcolor[cmyk]{0.8000000000000002,0.4000000000000001,0,0}\textcolor{white}{0.80} & \cellcolor[cmyk]{0.5545454545454546,0.2772727272727273,0,0}\textcolor{white}{0.55} \\
        DSTC10 & App. &  \cellcolor[cmyk]{0.880952380952381,0.4404761904761905,0,0}\textcolor{white}{0.88} & \cellcolor[cmyk]{0.4761904761904762,0.2380952380952381,0,0}\textcolor{white}{0.48} & \cellcolor[cmyk]{0.042857142857142864,0.28571428571428575,0.28571428571428575,0}\textcolor{white}{-0.29} & \cellcolor[cmyk]{0.0642857142857143,0.4285714285714286,0.4285714285714286,0}\textcolor{white}{-0.43} & \cellcolor[cmyk]{0.042857142857142864,0.28571428571428575,0.28571428571428575,0}\textcolor{white}{-0.29} & \cellcolor[cmyk]{0.9047619047619048,0.4523809523809524,0,0}\textcolor{white}{0.90} & \cellcolor[cmyk]{0.9047619047619048,0.4523809523809524,0,0}\textcolor{white}{0.90} & \cellcolor[cmyk]{0.5714285714285715,0.28571428571428575,0,0}\textcolor{white}{0.57} \\
               & Acc. &  \cellcolor[cmyk]{0.8095238095238096,0.4047619047619048,0,0}\textcolor{white}{0.81} & \cellcolor[cmyk]{0.5476190476190477,0.27380952380952384,0,0}\textcolor{white}{0.55} & \cellcolor[cmyk]{0.003571428571428572,0.023809523809523815,0.023809523809523815,0}\textcolor{black}{-0.02} & \cellcolor[cmyk]{0.021428571428571432,0.14285714285714288,0.14285714285714288,0}\textcolor{black}{-0.14} & \cellcolor[cmyk]{0.003571428571428572,0.023809523809523815,0.023809523809523815,0}\textcolor{black}{-0.02} & \cellcolor[cmyk]{0.880952380952381,0.4404761904761905,0,0}\textcolor{white}{0.88} & \cellcolor[cmyk]{0.8095238095238096,0.4047619047619048,0,0}\textcolor{white}{0.81} & \cellcolor[cmyk]{0.7857142857142858,0.3928571428571429,0,0}\textcolor{white}{0.79} \\
        \hline
    \end{tabular}
    \end{center}
\end{table*}

For evaluation, we use the same metrics as originally proposed by \citet{kimDomainAPIsTaskoriented2021} for this task.
The detection subtask is evaluated by the precision, recall, and F1 score of the predictions.
For the selection task, the ordered list of the top five retrieved documents is evaluated using recall at one (R@1), recall at five, and the mean reciprocal rank at five.
Finally, the generation task is evaluated using BLEU 1-4, METEOR, and ROUGE 1, 2, and L.
Since during inference, the results in the selection and generation subtask depend on the results of the previous subtasks,
 \citet{kimDomainAPIsTaskoriented2021} propose to re-weight the result by calculating the F1 score using the number of true positives, false positives, and false negatives from the detection task.
To give an overall rank the mean reciprocal rank of all automatic metrics is calculated.
Since eight metrics are used for the generation task but only three metrics for the detection and selection tasks, this gives a higher weight to the generation task.

In addition to these automatic metrics, a human evaluation was performed on the best entries of the finalists based on automatic scores.
Therefore, crowd workers were asked to score the appropriateness and the accuracy of each turn on a scale of 1 to 5.
The appropriateness indicates how well a system response is naturally connected to a given dialog context.
In particular, this means that it evaluates whether the generated response is a sensible response to the question asked by the user.
In addition, a response that is naturally connected to the dialog should be written in the same style as the rest of the dialog.
To evaluate this, crowd workers were only given the dialog context and the response but not the corresponding document to avoid that they are influenced by the correctness of the response.
For the accuracy score, crowd workers should judge how accurate a system response is given the provided reference document.
In addition to the generated response and the reference document, the judges are also provided with the dialog context.
Due to this, it is not clear whether just the accuracy of the response given the reference document was judged or whether the judges also considered whether the generated response is a valid response to the user's question.
The final ranking in the human evaluation is then calculated based on the average of the appropriateness and the accuracy score.

\begin{table*}
    \caption{Results on the DSTC9 test set.}
    \label{table:dstc9_results}
    \centering
    \begin{tabular}{l|r|r|r|r|r|r|r|r|r|r}
      \hline
       & detection   & selection     & \multicolumn{3}{l|}{generation} & \multicolumn{3}{l|}{factuality}         & \multicolumn{2}{l}{human evaluation} \\
                     & {F1}          & {R@1}  & {BLEU-1} & {METEOR} & {ROUGE-L} & BLEU-1 & F1 & {$Q^2$} & {Acc.} & {App.} \\ \hline \hline
  Baseline & 94.5 & 62.0 & 30.3 & 29.8 & 30.4 & - & - & - & 3.72 & 3.94\\ \hline\hline
  Team 19, Knover \cite{heLearningSelectExternal2021} & 98.9 & 92.3 & 38.0 & 38.7 & 37.4 & 34.6 & 56.6 & 69.1 & \textbf{4.39} & \textbf{4.39}\\ \hline
  Team 3 \cite{miGeneralizedModelsDomain2021} & \textbf{99.1} & 90.1 & 38.6 & 39.1 & \textbf{38.9} & 35.8 & 56.9 & 73.0 & 4.35 & 4.36\\ \hline
  Team 10 \cite{tangRADGERelevanceLearning2021} & 97.3 & 91.6 & 36.8 & 37.2 & 36.9 & 36.2 & 57.3 & 70.5 & 4.35 & 4.32\\ \hline
  Team 15 & 98.0 & 89.8 & 37.8 & 39.3 & 37.6 & \textbf{43.8} & \textbf{69.5} & \textbf{80.9} & 4.38 & 4.28\\ \hline
  Team 17 & 98.4 & 87.1 & 37.0 & 37.2 & 36.9 & 36.1 & 60.0 & 71.2 & 4.34 & 4.31\\ \hline
  Team 18 (our) \cite{thulkeEfficientRetrievalAugmented2021} & 96.4 & 89.9 & 37.9 & 38.6 & 37.1 & 32.1 & 53.9 & 66.7 & 4.33 & 4.29\\ \hline\hline
  \citet{jinCanBeFurther2021} & 98.7 & {92.5} & {35.8} & \textbf{43.8} & 35.4 & - & - & - & - & -\\ \hline
  This work & 98.8 & \textbf{92.9} & \textbf{39.3} & {39.8} & 38.5 & 33.1 & 55.0 & 69.5 & - & -\\ \hline\hline
  \multicolumn{9}{l|}{Ground-truth} & 4.59 & 4.45 \\ \hline
    \end{tabular}
  \end{table*}

The human judgments for the finalists in both challenges also allow us to evaluate the different automatic metrics regarding their correlations to human judgments.
Therefore, we calculate the Spearman rank correlation coefficient between all pairs of automatics metrics and human judgments over all finalists' entries.
The results for a subset of the metrics are shown in \Cref{tab:correlation}.
\citeauthor{kimDSTC10TaskOverview2022} did the same analysis in their summary papers for both challenges \cite{kimDomainAPIsTaskoriented2021, kimDSTC10TaskOverview2022}.
We note that the correlation coefficients for the DSTC9 tasks differ slightly from the ones reported by \citeauthor{kimDSTC10TaskOverview2022} since not all teams agreed that their generated responses were published.

For both challenges, we can observe that the detection and selection metrics are relatively well correlated to human judgments.
While for DSTC9 this was also the case for the generation metrics, there is no or even negative correlation of the generation metrics to human judgments for DSTC10.
An explanation for this is that all generation metrics are based on lexical overlap and thus favor responses with high lexical similarity to the reference.
This means for DSTC10 that participants that did not adapt the style of the generated responses to the spoken dialogs only achieved low scores in these metrics.
Given the current definition of appropriateness, it is not clear whether this also covers the style of the generated responses.
During the human evaluation for DSTC10, judges did not seem to have a preference for generated responses that tried to replicate the style of the rest of the dialog.
Given that, it is also an open question whether replicating the spoken style of a dialog in a system response is even desirable or whether a cleaner written response would be preferable.

In general, one can conclude that the current generation metrics are at least not suitable to evaluate how accurately the information from the selected document is reflected in the generated response.
To address this, recently multiple factuality metrics for knowledge-grounded dialog were proposed.
Most of these metrics evaluate the factuality of a response by comparing it to the reference document instead of the reference response.
\citet{bert-score} calculate the BLEU score between the generated response and the document. In this work, we calculate the BLEU-1 score as in the normal generation metrics.
\citet{dinanWizardWikipediaKnowledgePowered2018} propose to use the token-level F1 overlap of the generated response and the document.
Finally, as an alternative to metrics based on lexical overlap is the $Q^2$ metric proposed by \citet{honovich-etal-2021-q2}.
To calculate the $Q^2$ score first a model is used to identify answer candidates in the generated responses. Then a question generation model is used to generate questions for each answer candidate.
Next, a question answering model is used to extract answers to the generated questions from the reference document.
Finally, lexical overlap and a natural language inference model are used to calculate a score for how well the answer candidate from the response and the extracted span from the document match.
According to the evaluation by \citet{honovich-etal-2021-q2}, $Q^2$ has the highest correlation to human judgments for factuality of these three metrics.

We calculated all three of these metrics for the (public) generated responses of all finalists of DSTC9 and DSTC10 and show their correlation to human judgments in \Cref{tab:correlation}.
All three of them do correlate relatively well with accuracy for DSTC9 and DSTC10.
For DSTC9, as one would expect, there is no clear correlation to appropriateness since these metrics only consider the generated response and the document.
In contrast to the results by \citet{honovich-etal-2021-q2}, the lexical overlap-based metrics have a slightly higher correlation to human judgments than $Q^2$.
Similar to the original metrics, especially the lexical overlap-based metrics favor responses that have high lexical overlap with the reference document.
This causes the issue that responses that copy more from the document are favored which may result in less interesting results. Additionally, responses that are written in a similar style to the document may be favored.
This may also be undesirable since it penalizes systems that adapt the style of their responses to the style of the dialog.

\section{Results}

The experiments have been done using HuggingFace Transformers \cite{wolfTransformersStateoftheArtNatural2020} and Sisyphus \cite{peterSisyphusWorkflowManager2018}.
All models were trained on Nvidia GTX 1080 Ti or RTX 2080 Ti GPUs.
In the selection and generation subtasks which depend on the results of previous tasks, we evaluate the methods on the ground truth labels to facilitate comparability.

\subsection{DSTC9}

\Cref{table:dstc9_results} shows the baseline, our results, and the results of the top 5 teams in the DSTC9 evaluation.
According to the automatic evaluation, we achieved 6th and according to human evaluation 7th place out of in total 24 submissions in the challenge.
Our best system used Hierarchical Selection and Retrieval Augmented Generation.
For our official DSTC9 submission, we did not yet use any additional data augmentation or model ensembles.

The line labeled \enquote*{This work} shows our current results on the DSTC9 test set.
For the detection and selection, we apply the data augmentation discussed above.
For the generation, we use the noisy channel model instead of the RAG model.
For a more detailed analysis of the effect of the RAG and Noisy Channel Model, we refer readers to \citet{thulkeEfficientRetrievalAugmented2021} and \citet{noisyChannelPaper2022}.

\subsection{DSTC10}

\begin{table*}
  \caption{Results on the DSTC10 test set.}
  \label{table:dstc10_results}
  \centering
  \begin{tabular}{l|r|r|r|r|r|r|r|r|r|r}
    \hline
     & detection   & selection     & \multicolumn{3}{l|}{generation} & \multicolumn{3}{l|}{factuality}         & \multicolumn{2}{l}{human evaluation} \\
                   & {F1}          & {R@1}  & {BLEU-1} & {METEOR} & {ROUGE-L} & BLEU-1 & F1 & {$Q^2$} & {Acc.} & {App.} \\ \hline \hline
Baseline: DSTC9 & 79.5 & 45.8 & 11.5 & 12.2 & 11.4 & - & - & - & 2.74 & 2.79\\ \hline
Baseline: Knover \cite{heLearningSelectExternal2021} & 76.9 & 49.5 & 12.5 & 13.6 & 12.3 & - & - & - & 2.78 & 2.74\\ \hline\hline
Team B10 \cite{tian2021tod-da} & 92.3 & \textbf{79.3} & 16.2 & 21.0 & 21.9 & \textbf{47.5} & \textbf{62.9} & \textbf{77.7} & \textbf{3.49} & \textbf{3.35}\\ \hline
Team B04 \cite{yan2022generalized} & 91.8 & 74.8 & 33.8 & 40.7 & 38.7 & 16.3 & 27.7 & 66.4 & 3.34 & 3.30\\ \hline
Team B08 (our) \cite{thulkeAdaptingDocumentGrounded2022} & 91.1 & 71.0 & \textbf{40.1} & \textbf{46.0} & \textbf{44.0} & 15.9 & 26.7 & 68.8 & 3.34 & 3.26\\ \hline
Team B14 \cite{zhang2021knowledgegrounded} & \textbf{92.4} & 62.0 & 27.1 & 31.7 & 31.8 & 28.0 & 42.4 & 61.3 & 3.29 & 3.28\\ \hline
Team B02 & 90.4 & 69.3 & 37.3 & 43.9 & 41.1 & 14.4 & 23.7 & 66.1 & 3.29 & 3.23\\ \hline\hline
\multicolumn{9}{l|}{Ground-truth} & 3.58 & 3.48 \\ \hline
  \end{tabular}
\end{table*}

\Cref{table:dstc10_results} shows the official results of DSTC10 Track 2 Task 2 of the best five teams according to the human evaluation.
The baseline system is the original baseline system proposed by \citet{kimDomainAPIsTaskoriented2020} for DSTC9.
In total, 16 teams participated in the challenge.
Our best system achieved 4th place in the detection (F1) and selection (R@1) subtasks and 1st place in the generation subtask.
In the official ranking according to the automatic metrics, our system achieved 1st place and in the human evaluation, 3rd place.

\subsection{Ablation Analysis}

\subsubsection{Detection}

\begin{table}
    \caption{Effect of different text preprocessing techniques in the detection task on the DSTC10 validation data.}
    \label{table:text_norm}
\begin{center}
\begin{tabular}{l|r} 
    \hline
    method & F1 \\\hline 
    baseline (RoBERTa-large) & 75.3 \\ 
    + lowercasing   & 78.4 \\
    \hphantom{+ }+ no punct.    & 79.7 \\ 
    \hphantom{+ }\hphantom{+ }+ numbers written out    & 83.7 \\
    \hphantom{+ }\hphantom{+ }\hphantom{+ }+ no abbrev.    & 84.1 \\ 
\end{tabular}
\end{center}
\end{table}

\begin{table}
    \caption{F1 scores of the detection subtask on~the~DSTC10~validation~and~test~data.}
    \label{table:detection_results}
    \centering
    \begin{tabular}{l|r|r}
        \hline
        model & validation & test  \\ \hline
        baseline (+ text preprocessing) & 84.8 & 84.1 \\
        + data augmentation & 91.9 & 85.1 \\
        \hphantom{+ }+ in-domain pretraining & 93.5 & 86.0 \\
        \hphantom{+ }\hphantom{+ }+ ASR n-best (weighted) & 94.5 & 86.5 \\
        \hphantom{+ }\hphantom{+ }+ ASR n-best (max) & 94.7 & 87.7 \\
        \hphantom{+ }\hphantom{+ }\hphantom{+ }+ DSTC9 test + DSTC10 val & - & 90.5 \\
        \hphantom{+ }\hphantom{+ }\hphantom{+ }\hphantom{+ }+ ensemble & - & 91.1 \\
    \end{tabular}
\end{table}

For the DSTC9 task, as described above the largest improvements for the detection task come from additional data augmentation and using DeBERTa instead of RoBERTa.
We experimented with the different proposed text processing strategies for the detection task.
\Cref{table:text_norm} shows the results on the DSTC10 test data.
We observed that each method gives a slight improvement in the final performance.
Therefore, we decided to apply these pre-processing methods in the detection and selection tasks.

\Cref{table:detection_results} shows the results of our proposed methods for the detection task on the DSTC10 validation and test data.
First, augmenting the training data with additional samples generated from the knowledge base gave us a strong improvement on the validation and a small improvement on the test data.
The additional, in-domain pretraining of the RoBERTa model further improved the results by 1\%.
Next, we experimented with the two proposed ASR n-best strategies and observed better results with the max strategy.
Finally, we included the DSTC9 test and DSTC10 validation data into the training of the model and created an ensemble of different training runs.

For DSTC9, we compare the latency of our approach to the approach of the winning team (Team 19, Knover \cite{heLearningSelectExternal2021}) in \Cref{table:latencies}.
Due to their Schema Guided Knowledge Decision (SGKD) approach, they do a separate forward pass for each document and schema description.
Additionally, they fine-tune the 1.6B parameter PLATO-2 model \cite{bao-etal-2021-plato} which is around four times larger than the models that we are using.

\subsubsection{Selection}

\begin{table}
    \centering
    \caption{Selection results on the DSTC9 validation and test data.}
    \label{table:selection}
	\begin{tabular}{l|rr|rr}
		\hline
										 & \multicolumn{2}{l|}{validation} & \multicolumn{2}{l}{test} \\ 
				 				  		 & R@1 & R@5 & R@1 & R@5 \\ \hline \hline
		baseline (ours)            		 & 91.9 & \textbf{99.7} & 91.0 & \textbf{99.3} \\ \cline{2-5}
		 - w/o long context			     & 81.4 & 97.9 & 84.0 & 97.7\\ \cline{2-5}
		 - w/o domain in input		         & 71.8 & 92.8 & 65.2 & 81.4 \\ \hline
		Hierarchical$_\text{domain+entity,doc}$& 96.3 & 98.1 & \textbf{93.2} & 97.3 \\ \cline{2-5}
		Hierarchical$_\text{domain,entity,doc}$& \textbf{96.8} & 98.6 & 88.1 & 91.1 \\ \cline{2-5}
		Bi-Encoder Triplet               		 & 90.9 & 98.9 & 87.2 & 96.9 \\ \cline{2-5}
		 - w/o RAG				      		 & 85.6 & 97.0 & 82.6 & 93.3 \\ \hline
        Bi-Encoder Triplet hard          		 & 90.8 & 98.9 & 83.8 & 95.2 \\ \cline{2-5}
		 - w/o RAG					 		 & 88.0 & 97.6 & 84.7 & 95.8 \\ \hline
        Bi-Encoder NLL                   		 & 93.4 & 98.8 & 85.5 & 96.8  \\ \cline{2-5}
        - w/o RAG          		 		 & 90.1 & 98.1 & 84.0 & 94.4 \\ \hline
        Bi-Encoder NT-Xent (w/o RAG) & 95.8 & 99.6 & 89.0 & 98.3 \\ \hline
	\end{tabular}
\end{table}

\Cref{table:selection} compares the different methods we proposed for the selection task on the DSTC9 validation and test data.
The \emph{Hierarchical Selection} model achieves the best results for MRR@5 and R@1 of all selection models.
Other models outperform the model concerning R@5.
One explanation is that the model only returns documents of a single entity in its final ranking, thus these numbers are not fully comparable.
When analyzing the improvements, we mainly see that the number of confusions among similar documents of different entities reduces.
The model first decides which domain and entity are relevant before selecting the document.
Furthermore, Hierarchical Selection generalizes very well to new domains and sources (R@1 of 98.5 for attraction, 94.4 for sf\_written, and 87.5 for sf\_spoken).
As expected, it achieves a significant speedup of 20x compared to the baseline selection method as shown in \Cref{table:latencies}.
Even so, for a real-time application, a latency of around 13 seconds is still too high.
Team 19, Knover, uses the baseline approach for selection and a 1.6B parameter models and thus is even slower than our baseline.

The bi-encoder model achieves an additional speedup of more than 100x compared to the hierarchical selection model and more than 2,500x compared to the baseline model.
On the validation data, we observed that the negative log-likelihood (NLL) loss outperforms the triplet loss and even achieves better results than the baseline method.
Nevertheless, the model trained using the triplet loss seems to generalize better to the test data where it outperforms the model trained using the NLL loss.
As shown in \Cref{table:selection}, the performance of the models is significantly improved by joint training with the RAG model.
One interesting observation is that the bi-encoder models do not generalize well to the spoken data (R@1 goes down to 43.2 for the bi-encoder NLL model) compared to other models.
Finally, as described above we also train a bi-encoder using the NT-Xent loss with in-batch negatives.
We use a batch size of 64 and a temperature value of 20.
Using this approach we outperform even the bi-encoder models we previously fine-tuned with RAG.


\begin{table}
    \caption{Runtimes in seconds per turn for different methods on~one~GTX~1080~Ti with batch size 1.}
    \label{table:latencies}
    \centering
	\begin{tabular}{l|l|r|r}
	\hline
	task       & model 				 & \multicolumn{2}{c}{runtime} \\
		   	   &				 	 & validation	   & test   \\ \hline
	detection  & baseline 		 & 0.04    & 0.04   \\
    & Team 19, Knover \cite{heLearningSelectExternal2021} & 688.16 & 2,790.13 \\
    & - w/o SGKD \cite{heLearningSelectExternal2021} & 0.23 & 0.23 \\ \hline
	selection  & Cross-Encoder 		 & 111.66  & 276.53 \\
		   	   & Hierarchical		 & 4.60    & 13.79  \\
	   	       & Bi-Encoder				 & 0.04    & 0.04   \\
               & Team 19, Knover \cite{heLearningSelectExternal2021}  &658,73 & 2.734,63 \\ \hline
	generation & baseline			 & 0.85    & 0.82   \\
		       & RAG + Bi-Encoder 			 & 1.20    & 1.48   \\
               & Team 19, Knover \cite{heLearningSelectExternal2021}   &1.74 & 1.74 \\ \hline

	\end{tabular}
\end{table}

\begin{table}
    \caption{R@1 scores of the selection subtask on~the~DSTC10~validation~and~test~data.}
    \label{table:selection_results}
    \centering
    \begin{tabular}{l|r|r}
        \hline
        model & validation & test  \\ \hline
        baseline (+ text preprocessing) & 71.2 & 70.0 \\
        + Beam Search & 74.0 & 73.5 \\
        \hphantom{+ }+ Taskmaster \& DSTC10 data & 78.8 & 76.3 \\
        \hphantom{+ }\hphantom{+ }+ in-domain pretraining & 83.7 & 77.0 \\
        \hphantom{+ }\hphantom{+ }\hphantom{+ }+ ASR n-best (max) & 79.8 & 77.7 \\
        \hphantom{+ }\hphantom{+ }\hphantom{+ }+ ASR n-best (weighted) & 81.7 & 77.7 \\
        \hphantom{+ }\hphantom{+ }\hphantom{+ }\hphantom{+ }+ DSTC9 test + DSTC10 val & - & 77.3 \\
        \hphantom{+ }\hphantom{+ }\hphantom{+ }\hphantom{+ }\hphantom{+ }+ ensemble & - & 77.6  \\
    \end{tabular}
\end{table}

\Cref{table:selection_results} shows the results of applying our proposed methods for the selection task on the DSTC10 validation and test data.
Using our proposed Beam Search approach instead of always taking the entity with the highest score results in an improvement of around 3\% absolute.
Further, training the domain and entity selection model on additional data from Taskmaster and DSTC10 Task 1 gives an additional improvement.
On the validation data, we observed slight degradations with our strategies to handle the ASR n-best list.
On the test set, this resulted in improvements.
We assume that the observed degradations can be attributed to the small size of the validation set.
In contrast to the detection task, including the DSTC9 test and DSTC10 validation data resulted in small degradations.
Finally, an ensemble of different training runs slightly improved the results again.

\section{Related Work}

We review different methods introduced for DSTC9 for the tasks of Turn Detection, Knowledge Selection, and Response Generation.
\citet{heLearningSelectExternal2021} propose to model the first task by deciding whether a knowledge document or schema description obtained from MultiWOZ is more likely to be sought by the user.
In the first case, the most likely knowledge document is selected and in the latter case, the turn is deemed not knowledge-seeking.
Furthermore, similar to \citet{jinCanBeFurther2021} the authors propose different strategies to sample negative training examples, such as sampling documents from the same domain or entity.
\citet{tangRADGERelevanceLearning2021} propose to select negatives by first training a model on the selection task with negatives sampled from the same entity or domain and then taking the documents likely to be confused under the model as negatives to train a stronger model.
While this forms an explicit negative sampling, \citet{thulkeEfficientRetrievalAugmented2021} explore to fine-tune the selection models end-to-end with the response generation task by using a retrieval augmented model \cite{guuREALMRetrievalAugmentedLanguage2020, lewisRetrievalAugmentedGenerationKnowledgeIntensive2020}, where the marginalization can be seen as an implicit batching of hard negatives.
Furthermore, the authors propose to use a hierarchical selection approach and formulate knowledge selection as a metric learning problem using bi-encoders, similar to \citet{karpukhinDensePassageRetrieval2020}.
Finally, \citet{miGeneralizedModelsDomain2021} and \citet{kimHowRobustEvaluating2021} propose different data augmentation methods to augment the training data by unseen knowledge documents.


The Noisy Channel decomposition \cite{shannonnoisychannel} has been widely used in different language technology tasks, such as machine translation \cite{brown-etal-1993-mathematics} or automatic speech recognition \cite{bahl_jelinek}.
With the advent of deep learning, modeling these tasks discriminatively has often been the preferred choice.
Nevertheless, recently neural noisy channel modeling has been explored for different tasks, such as neural machine translation \cite{yuNeuralNoisyChannel2017, yeeSimpleEffectiveNoisy2019, jeanLogLinearReformulationNoisy2020, subramanian2021nvidia}, few-shot text classification \cite{minNoisyChannelLanguage2021}, and dialog \cite{liuPretrainingNoisyChannel2021a}.

\section{Conclusion and Future Work}
In this work, we summarize our submissions to the DSTC9 and DSTC10 challenges.
In addition to that, we applied our recent findings on the DSTC9 test set and achieved state-of-the-art results on multiple automatic metrics.
We also revisited the bi-encoder model and showed that with recent advances in training these models we were able to reduce the gap to cross-encoders for this task.
It would be interesting to see whether this gap can be closed even further.
Finally, we explored alternatives for automatic metrics for the generation task by exploring a few recently proposed factuality metrics.
While these metrics also are not ideal, we showed that they at least better align with human judgments.

Kim et al. already announced a third iteration of the task for the 11th Dialog System Technology Challenge (DSTC11).
In contrast to the previous iterations, the documents in the knowledge base will include subjective user reviews and more than one document may be relevant for a user turn.
Under these conditions, generation approaches that are capable of grounding the responses in multiple documents, like the proposed RAG approach, presumably become crucial.
\section*{Acknowledgements}
This work was partially supported by the project HYKIST funded by the German Federal Ministry of Health on the basis of a decision of the German Federal Parliament (Bundestag) under funding ID ZMVI1-2520DAT04A, and by NeuroSys which, as part of the initiative \enquote{Clusters4Future}, is funded by the Federal Ministry of Education and Research BMBF (03ZU1106DA).

\bibliography{\jobname, david, nico, manual}



 





\end{document}